\documentclass[11pt]{article}
\usepackage{konvens2018}
\usepackage{mathptmx}
\usepackage[scaled=.90]{helvet}
\usepackage{courier}
\usepackage{graphicx}
\usepackage{booktabs}
\usepackage{tabularx}

\usepackage{url}
\usepackage{latexsym}

\usepackage{todonotes}

\title{microNER: A Micro-Service for German Named Entity Recognition \\ 
based on BiLSTM-CRF}

\author{Gregor Wiedemann \qquad Raghav Jindal \qquad Chris Biemann \\
Language Technology Group\\ 
Department of Informatics\\ 
Universit\"{a}t Hamburg, Germany\\
 {\tt \{gwiedemann, biemann\}@informatik.uni-hamburg.de} \\
 {\tt raghavjindal2003@gmail.com}}

\date{}

\begin{document}
\maketitle
\begin{abstract}

For named entity recognition (NER), bidirectional recurrent neural networks became the state-of-the-art technology in recent years. Competing approaches vary with respect to pre-trained word embeddings as well as models for character embeddings to represent sequence information most effectively. 
For NER in German language texts, these model variations have not been studied extensively. We evaluate the performance of different word and character embeddings on two standard German datasets and with a special focus on out-of-vocabulary words.
With F-Scores above 82\% for the GermEval'14 dataset and above 85\% for the CoNLL'03 dataset, we achieve (near) state-of-the-art performance for this task.
We publish several pre-trained models wrapped into a micro-service based on Docker to allow for easy integration of German NER into other applications via a JSON API.
\end{abstract}

\section{Introduction}
In information extraction, named entity recognition (NER) is the task to automatically identify proper nouns in natural language texts and classify them into predefined categories such as person, location or organization. Usually, it is approached as a sequence classification task. For the English language, the problem has been well studied and solved by current state-of-the-art neural network models achieving very high levels of accuracy.

For documents in German language, NER performance has traditionally been much lower due to more complex linguistic structures such as compounds, separation of verb prefixes and the use of uppercase letters not only to indicate proper nouns but also regular nouns. At the latest shared task event on German NER,  GermEval 2014 \cite{Benikova.2014}, only two neural network-based systems took part in the competition. At that time they were not able to outperform the winning team's submission that relied solely on Conditional Random Field (CRF) models and heavy use of external linguistic resources (POS-tags, semantic word clusters, and gazetteers compiled from Wikipedia, OpenStreetMap, and other databases).

Since then, a number of innovations from natural language processing (NLP) have been introduced to sequence classification using deep neural networks. In this paper, we utilize recurrent neural models for sequence classification in German texts based on previous work for English NER. 
The major advantage of neural models compared to previous freely available German NER systems such as \newcite{Benikova.2015} is that they do not rely on any external linguistic resources other than word embeddings. This drastically reduces the effort in feature engineering. 

In recent years, bidirectional recurrent neural networks (RNN) combined with CRF became the de-facto standard model for sequence classification tasks.
Competing approaches vary these models with respect to pre-trained word embeddings as well as models for character embeddings. The main challenge approached by complex embeddings is to represent sequence information most effectively not only with respect to the training data but also in generalization to data unseen during training. 

To study the effects of character and word embeddings on German NER, we compare different variants of the standard RNN approach. We evaluate the different models on two standard datasets for German NER. For our best model, we can report a new state-of-the-art for one of the datasets, and a second best result for the other dataset.
We publish pre-trained models wrapped into a micro-service based on Docker to allow for easy integration of NER into external applications.

The paper is structured as follows: In Section~\ref{sec:related}, we introduce related work on the current state of the art in NER with a special focus on the results for the German language. Then, we describe our own base model for the task together with a number of variants for learning character embeddings in Section~\ref{sec:model}. An extensive evaluation of these variations of the base model architecture is presented in Section \ref{sec:eval}. In addition to character embeddings, we further study how different word embedding models affect the results with a special focus on out-of-vocabulary words. The best final models for NER are proliferated as a ``micro-service'' described in Section~\ref{sec:microservice}, before we conclude with a brief discussion of our results with respect to the most recent developments in NLP (Section~\ref{sec:discussion}).

\section{Related Work}
\label{sec:related}

Most of the progress in NER has been made over the last years by the use of recurrent neural network architectures such as Long Short-Term Memory (LSTM) \cite{Hochreiter.1997} for sequence classification on the one hand and pre-trained word embeddings to encode semantics of token sequences on the other hand.
One decisive problem in sequence classification is to represent information for out-of-vocabulary (OOV) words in the model. For practical applications, especially for NER, it cannot be expected that all entities have already been seen during training, neither for NER nor during training of word embeddings. 

Information for OOV words in CRF models is typically captured by context and word shape features as well as word clusters learned from unlabeled training sets. With such a model, \newcite{Haenig.2014} achieved 76.4~\% F1-score on the Germ\-Eval 2014 dataset.
For neural network architectures, \newcite{Dossantos.2015} first introduced character embeddings combined with 1D-convolution (CNN) to learn sub-word representations for OOV words, which drastically improved NER performance. 

\newcite{Chiu.2016} extended this model by using bidirectional LSTM layers instead of a word level classification. 
\newcite{Hovy.2016} added a CRF-Layer on top of this. 
Since then, bidirectional LSTMs combined with CRF became the de-facto standard model for sequence classification.  
Based on this, \newcite{Lample.2016} suggested a novel architecture for learning character embeddings. Instead of a CNN layer, they use a second BiLSTM layer for character sequences from words. 
All of the last three models achieve F1-scores above 91\% for English datasets. 

It is a widely known fact in the field of machine learning that differences in the performance of neural network models cannot be obtained from single runs only but must be inferred from averages of multiple runs with different random seeds. Unfortunately, not all studies report clearly about this. This makes it hard to decide which model actually constitutes the current state of the art. According to \newcite{Reimers.2017}, \newcite{Lample.2016} produce on average most accurate NER results for English although they do not report the highest F-score.

For German NER, there are much less comparative studies. The results from GermEval 2014 have been slightly topped by \newcite{Agerri.2016} who used a Perceptron model in combination with multiple word clustering. 
For a longer time, the best result for the German CoNLL 2003 dataset was also reported by \newcite{Lample.2016}.
Explicitly for German NER, \newcite{Riedl.2018} recently presented an evaluation of the standard neural model on four different German datasets comparing it to conventional CRF classification. They conclude that BiLSTM-CRF architectures in general outperform CRFs, and are especially suited to profit from transfer learning. This makes them attractive also in scenarios in which only little training data is available. Most recently, a new state of the art for many sequence labeling tasks, among others German NER, has been achieved by \newcite{Akbik.2018}. They employ a simple BiLSTM-CRF network together with specifically trained contextual word embeddings that allow disambiguation of homonymous terms in a sentence. 

\begin{figure}[h]
\centering
\hspace*{-0.5cm}
\includegraphics[width=0.52\textwidth]{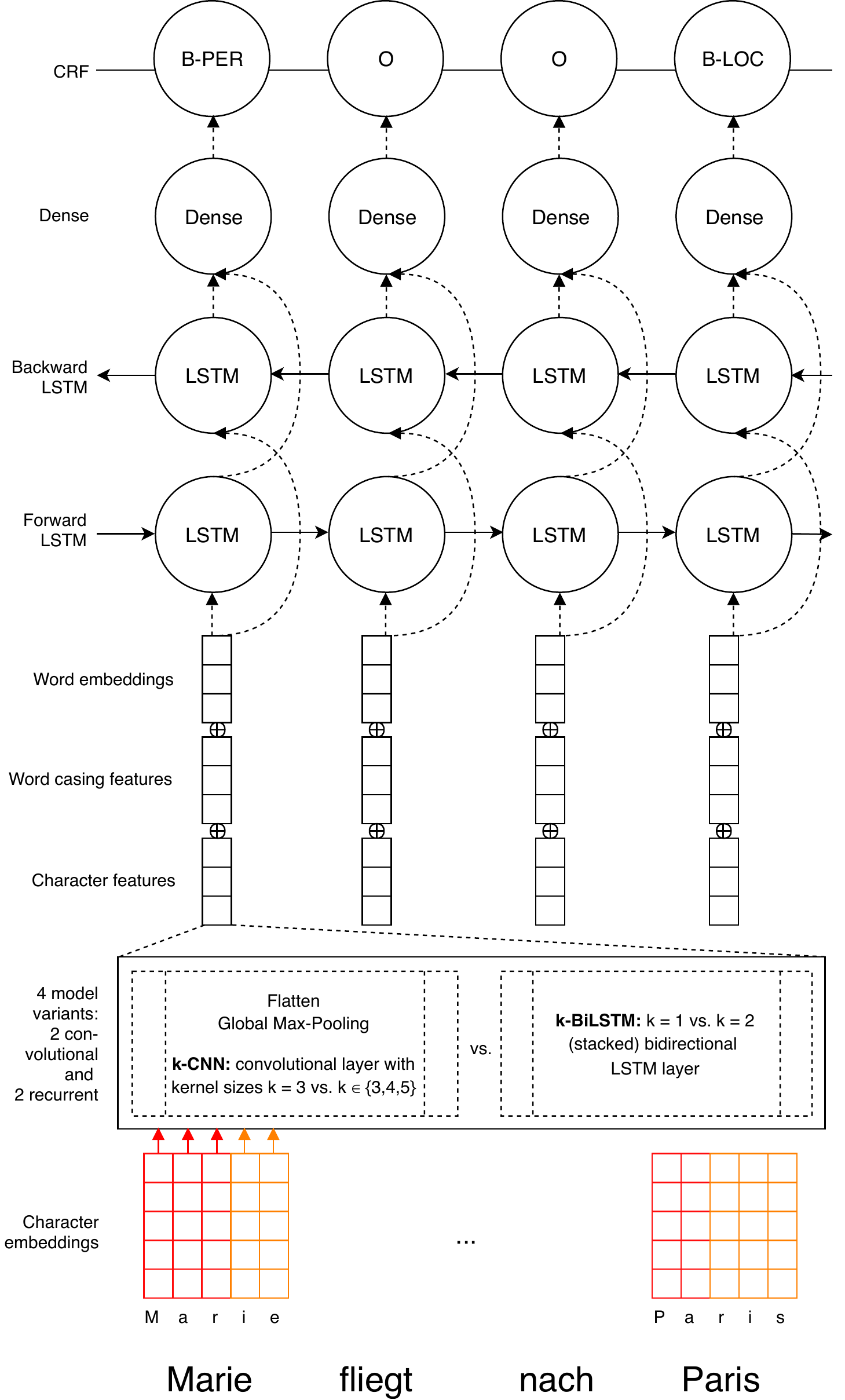}
\caption{BiLSTM-CRF architecture with four variants to model character embeddings.} 
\label{fig:architecture}
\end{figure}

\section{Variants of BiLSTM-CRF for NER}
\label{sec:model}

\paragraph{Base model:} We build our model for German NER based on BiLSTM and CRF by combining elements from the architectures used in \newcite{Chiu.2016} and \newcite{Lample.2016}.\footnote{For detailed mathematical descriptions of the different layers of the neural network model (i.e. BiLSTM, CRF and CNN) see the referenced papers.} The overall model architecture is displayed in Figure~\ref{fig:architecture}.
The main component of the network is a BiLSTM layer for the token sequence of a given sentence consisting of 200 cells, followed by a linearly activated dense layer with as many units as there a distinct sequence labels in the training set. Finally, the most likely label sequence is obtained from a CRF layer. The BiLSTM layer is fed with three different types of features to learn from: 1. word embeddings of 300 dimensions in length (fixed during training), 2. seven binary casing features similar to those proposed by \newcite{Chiu.2016} (all lower, all upper, initial upper, numeric, mainly numeric, contains digit, other), and 
3. character embeddings of 32 dimensions trained by a second network architecture (see below).
As input for the BiLSTM layer, features of all three types are concatenated into one single vector. For regularization, we use input dropout and recurrent dropout at the BiLSTM layer, with each dropout rate set to 0.5. 


\paragraph{Character Embeddings:} Approaches in related works for English NER mainly differ in the way how character embedding features are learned. For this, \newcite{Lample.2016} argue that BiLSTM might be preferred over convolutional nets as used by \newcite{Hovy.2016} and \newcite{Chiu.2016}. Since recurrent networks are designed for sequential data such as character sequences they are better prepared to encode important prefix/suffix information from words. However, CNN architectures have been proven to be advantageous for many text classification tasks, and prefix/suffix information could also be encoded easily for them. 
For further investigation of this, we compare CNN and Bi\-LSTM layers for learning character embeddings in our experiments.

To represent prefix/suffix information for CNN explicitly, we introduce virtual characters \texttt{<S>} and \texttt{</S>} for sentence beginnings and endings, and \texttt{<W>} and \texttt{</W>} respectively for words. These virtual characters are pre-/appended to each character sequence obtained from single words to indicate beginnings and endings in the sequential character stream fed into the convolutional network.
We test two CNN setups, one with a single convolutional layer (\textit{CNN}), and one with three parallel convolutional layers (\textit{3-CNN}), each with filter size 32 and ReLu-activation. To learn the embeddings, the single CNN uses a kernel size of three characters. In the 3-CNN variant, each CNN has a different kernel size $s\in\{3,4,5\}$.
Convolutional filters are each followed by a global max-pooling layer, which is finally flattened to serve as character feature vector. 

For the BiLSTM-based character embedding model, we compare two variants as well. Since RNNs are already suited for sequential information, we do not feed  word and sentence beginnings/endings explicitly by virtual characters into the model. Instead, we just feed sequences of characters from single words  (pre-)padded to the same lengths into either a single BiLSTM layer (\textit{BiLSTM}), or into a stack of two BiLSTM layers (\textit{2-BiLSTM}), each with 50 LSTM cells per direction.

\paragraph{Learning:} The four model variations for character embeddings are compared to the base model without any character embeddings at all. We further compare how different pre-trained word embeddings contribute to NER.
Training is performed in mini-batches with Nesterov Adam optimization in two stages. In stage one, we train for up to 10 epochs with batch size 16. Model performance during training is evaluated by the F1-score of correctly classified NE chunks. The best classification model from this stage with respect to the validation set is used in stage two. In this stage, we train again for another 10 epochs with batch size 512. The best performing model from this stage is used as final model to determine the performance on the test set.

\section{Evaluation}
\label{sec:eval}

\subsection{Datasets} 

We evaluate the performance of our NER models on the two standard datasets that are available for the German language: the GermEval 2014 Shared Task dataset \cite{Benikova.2014}, and the CoNLL-2003 Shared Task dataset \cite{conll2003}. For both, we use the official evaluation script published together with the respective dataset.

\paragraph{GermEval'14:}
The  dataset contains 24,000 training sentences annotated with four main classes of entities: person (\texttt{PER}), location (\texttt{LOC}), organization (\texttt{ORG}) and other (\texttt{OTH}). For each class, two sub-classes \texttt{-deriv} and \texttt{-part} exist for entities that are either derived from an entity (e.g. `d\"anisch', \textit{Danish}) or belonging to a larger type (e.g. `Troja-Ausstellung', \textit{Troy exhibition}). Thus, in total the GermEval dataset contains 12 classes annotated in the BIO tagging schema. Further, two levels of annotation are provided. The first comprises outer chunks of most lengthy entities in a sentence, the second comprises inner chunks of nested entities such as in \texttt{[Real~[Madrid]\textsubscript{LOC}]\textsubscript{ORG}}. The official comparison metric for GermEval combines F1-scores for both outer and inner chunks. Since nested entities are rather rare, we concentrate on the classification of outer chunks for model comparison. But to be able to compare our approach to previous work, we also compute a second model for nested named entities and obtain the official score from the combined results.

\paragraph{CoNLL'03:}
The dataset \cite{conll2003} contains around 12,152 training sentences, also with four main classes for entities: person (\texttt{PER}), location (\texttt{LOC}), organization (\texttt{ORG}) and miscellaneous (\texttt{MISC}). The dataset was originally distributed with the IOB tagging schema. For coherence with the other dataset, we converted it also to the BIO tagging schema.
Although these main classes are quite similar to the GermEval data, there are no \texttt{-deriv} and \texttt{-part} sub-classes. Instances of GermEval's \texttt{-deriv} sub-classes often fit into the MISC-class of the CoNLL dataset. For the \texttt{-part} sub-classes there is no definite equivalent.

\subsection{Results} 

Most of the studies which contributed to the progress in NER over the last years did not evaluate their models on German data. To fill this gap, we test the de-facto standard BiLSTM-CRF neural model architecture with four different variations for character embeddings for German NER. We further test different, publicly available word embedding models. To check for the stability of the model performance between repeated rounds of learning, we run each experiment 10 times and report average results. 

\begin{table}
\centering
\caption{Performance of character and word embedding models (F1~\%)}
\label{tab:ablation}
\resizebox{0.48\textwidth}{!}{%
\begin{tabular}{lrr} \toprule
\textbf{Char embeddings}                  & \textbf{CoNLL}  & \textbf{GermEval} \\ 
\midrule
None        & 81.57        & 79.60              \\
CNN         & 81.78        & 80.17              \\
3-CNN       & 82.74        & 81.97              \\ 
BiLSTM      & \textbf{85.19}        & 82.12              \\ 
2-BiLSTM    & 84.87        & \textbf{82.19}              \\ \midrule
\textbf{Word embeddings} & \textbf{CoNLL} & \textbf{GermEval} \\ \midrule
Word2Vec               & 80.27       & 79.69 \\
Word2Vecf              & 82.13       & 80.32 \\
GloVe                  & 79.44       & 78.93 \\
fastText               & \textbf{82.74}       & \textbf{81.97} \\ \bottomrule
\end{tabular}}
\end{table}

\paragraph{Character embeddings:}
We start with the most basic setup, a BiLSTM network fed with fastText word embeddings and casing features but without any character embeddings (\textit{None}).
The next two setups introduce character embeddings based on convolutional network layers, one with a single convolutional layer with a kernel size of three characters (\textit{CNN}), a second combining three convolutional layers with different kernel sizes (\textit{3-CNN}).
The final two setups use recurrent layers, one single bidirectional LSTM layer (\textit{BiLSTM}), and two stacked layers (\textit{2-BiLSTM}).

Table~\ref{tab:ablation} shows that architectures modeling sub-word information with any character embedding approach improves the learning performance. 
Modeling sub-word information with more CNN layers leads to substantial performance increases. But, the highest performance gains are achieved by the recurrent models. Here, stacking more BiLSTM layers on top of each other does slightly improve the results on the GermEval dataset, although this performance gain is not significant. For the CoNLL dataset, the single BiLSTM layer performs best.

\begin{table}
\centering
\caption{In/out-of-vocabulary word performance}
\label{tab:oov}
\begin{tabularx}{0.48\textwidth}{Xllll} \toprule
\textbf{Model}                  & \multicolumn{2}{l}{\textbf{OOV}}  & \multicolumn{2}{l}{\textbf{IV}} \\ 
   & F1 (\%) & N & F1 (\%) & N \\ 
 \midrule
Word2Vecf              & 75.39  & 2421 & 86.88   & 2678\\
fastText               & 79.00  & 2895 & 85.48   & 2204 \\
\bottomrule
\end{tabularx}
\end{table}

\paragraph{\textbf{Word Embeddings:}}
The quality of pre-trained word embeddings is very decisive for the quality of NER results. \newcite{Reimers.2017} showed that for English datasets, GloVe embeddings provided by \newcite{Pennington.2014} outperform all other embeddings. But, the quality of the used GloVe embeddings is not only based on the model itself. It is rather determined by the fact that it was trained on several billion tokens of the Common Crawl corpus.\footnote{\url{http://commoncrawl.org}} 

To evaluate the contribution of different German word embedding models to the overall NER performance, we compare four different approaches based on the 3-CNN architecture:  \textit{word2vec} \cite{Mikolov.2013}, \textit{word2vecf} \cite{Levy.2014}, \textit{GloVe} \cite{Pennington.2014} and \textit{fastText} \cite{Bojanowski.2017}.\footnote{For fastText, we use the publicly available German model from \newcite{Bojanowski.2017} pre-trained on Wikipedia. For the other three, we trained our own models on German Wikipedia texts.}

The best performance is achieved using fastText embeddings. We suspect one main reason: due to its model architecture of determining embeddings not only from word contexts but also from sub-word information (here character n-grams from 3 to 6), we can determine word vectors also for OOV words, i.e. words that have not been part of the training data of the embedding model.

Although not capturing any sub-word information, the word2vecf model is achieving good performance, too. Word2vecf uses information from dependency parses to filter the context of words prior to learning their semantic embedding. As a result, syntactical aspects of similar word usage are much better covered by the model. This has already been shown to be beneficial for sequence-labeling tasks such as POS-tagging \cite{Kohn.2015}. Apparently, it also contributes positively to NER.

For a more detailed look at how the two best performing embedding models contribute distinctively to NER, we split the test set in two distinct subsets. One contains only those sentences whose every word is in the vocabulary of the embedding model; the other contains the remaining sentences, i.e. sentences with at least one OOV word. Table~\ref{tab:oov} shows the sizes of the test set split and their respective performance. 
While word2vecf performs slightly better on in-vocabulary words (IV) profiting from the dependency parse information during its learning phase, fastText actually performs way better on OOV words due to its subword information which allows inferring embeddings also for new unseen words. 
We assume that this is a favorable property when NER needs to be applied to text collections with language characteristics that differ from Wikipedia.

\begin{table*}
\centering
\caption{German NER results precision, recall, F1 \% (*~average of 10 runs for our model)}
\label{tab:finalresults}
\resizebox{\textwidth}{!}{%
\begin{tabular}{@{}llllllllll@{}}
\toprule
\textbf{Model}     & \multicolumn{3}{l}{\textbf{CoNLL*}} & \multicolumn{3}{l}{\textbf{GermEval (official)}} & \multicolumn{3}{l}{\textbf{GermEval (outer)*}} \\ 
          & P       & R      & F1     & P           & R           & F1          & P           & R          & F1         \\ \midrule
\newcite{Haenig.2014}     & -       & -      & -      & 78.07       & 74.75       & 76.38       & 80.67       & 77.55      & 79.08      \\
\newcite{Lample.2016}    & -       & -      & 78.76  & -           & -           & -           & -           & -          & -          \\
\newcite{Agerri.2016}    & 83.72   & 70.30  & 76.42  & 80.28       & 72.93       & 76.43       & 81.52       & 75.54      & 78.42      \\
\newcite{Riedl.2018}     & 87.67   & 78.79  & 82.99  & 81.95       & 78.13       & 79.99       & 83.07       & 80.62      & 81.83      \\
\newcite{Akbik.2018}   & -       & -      & \textbf{88.32}  & -           & -           & -           & -           & -          & -          \\ \midrule
Our best model & 87.11   & 83.36  & 85.19  & 81.50       & 80.17       & \textbf{80.83}       & 82.50       & 81.89      & \textbf{82.19}      \\ \bottomrule
\end{tabular}%
}
\end{table*}

\paragraph{Final results:}
Table~\ref{tab:finalresults} displays the results of our best performing models together with the reported results of previously published NER systems.
reported evaluation scores are the average of 10 runs for the CoNLL and GermEval (outer) dataset.\footnote{Since the official GermEval script combines scores for outer and (rarely occurring) nested NE chunks from two models, we report only the result from a single run of the nested NEs here.}
For the GermEval'14 dataset, we can report a new state of the art. For the CoNLL dataset, we achieve second highest scores ever reported. 
But, contextual embeddings introduced most recently by \newcite{Akbik.2018} seem to drastically improve results for many sequence classification tasks paving the way for a new technology trend in NLP.

\section{Micro-Service}
\label{sec:microservice}

NER is usually not an end in itself but one important step in a sequence of NLP-based text analysis steps. For this, we make pre-trained models based on the GermEval'14 and CoNLL'03 datasets available on our Github page under a free MIT license.\footnote{\url{https://uhh-lt.github.io/microNER}} 
The best performing models of the BiLSTM-CRF architecture with BiLSTM-based character embeddings and fastText word embeddings are selected for our micro-service.

We provide models trained individually on the CoNLL dataset, and on the GermEval dataset for outer as well as nested NEs. 
Further, we provide a model trained on both, CoNLL and GermEval outer NE chunks. 
Since most of GermEval's annotations from the \texttt{`-deriv'} sub-classes fit into CoNLL's \texttt{MISC} category, we map them to the \texttt{MISC} category for our combined model. The \texttt{`-part'} sub-class will be ignored in the combined model.
Table~\ref{tab:microservice} displays the overall performance of these models as well as with respect to their individual main classes. Performance scores are calculated on the respective (combined) test sets.

Moreover, for an easy integration into other applications, we publish the models wrapped into a Docker-based\footnote{\url{http://www.docker.com}} micro-service. The sequence classifier with our pre-trained models is wrapped in a web service based on the Flask Python framework,\footnote{\url{http://flask.pocoo.org}} which provides a simple RESTful JSON-API to exchange tokenized sentences and classified NE labels. This web service is built into a Docker container such that it can be easily deployed with a single \texttt{docker pull} command into any system having the Docker virtualization technology installed. Due to the micro-service architecture, which communicates over HTTP, our NER docker container can be easily be used in parallel NLP processing chains. We use it, for instance, in the information extraction pipeline of our ``new/s/leak'' project \cite{Wiedemann.2018g}, to create visualizations of co-occurrence networks of named entities.

\begin{table*}
\centering
\caption{Microservice model performance F1 \%}
\label{tab:microservice}
\begin{tabular}{@{}lrrrrr@{}}
\toprule
\textbf{Model}   & \textbf{PER} & \textbf{LOC} & \textbf{ORG} & \textbf{MISC} & \textbf{All labels} \\ \midrule
GermEval (outer) & 90.79        & 88.20        & 77.37        & 64.73         & 82.80        \\
GermEval (inner) & 29.46        & 63.53        & 22.95        & 0.00          & 55.91        \\
CoNLL            & 92.12        & 87.14        & 76.99        & 57.31         & 85.48        \\
Germeval+CoNLL   & 91.70        & 87.65        & 76.86        & 67.67         & 83.51        \\ \bottomrule
\end{tabular}
\end{table*}

\section{Discussion}
\label{sec:discussion}

We presented a comparative study of a BiLSTM-CRF base model for NER on German texts in combination with a variety of character and word embedding techniques. Recurrent BiLSTM networks to learn character embeddings together with fastText word embeddings have been proven most useful to capture information for words unseen during training resulting in significant improvements compared to the results from the last big Shared Task event for German NER.
For the first time, we can report an official F-score metric significantly above 80~\% for the GermEval 2014 dataset.
We publish a range of pre-trained models as a freely available Docker-based micro-service ready to use for other projects.

Recent works in NLP focus on contextual embeddings which are able to learn word embedding representations not only globally for a given corpus but in dependency of the concrete individual surrounding of a given context, e.g. a sentence. For NER, such contextual embeddings appear to be especially useful since they allow for a distinct representation of homonymous terms. In a sentence such as ``Von Jahr zu Jahr werden mehr Schiffe auf Kiel gelegt'', the embedding of the term `Kiel' would be dragged away from  German location names in the embedding space to a different semantic region containing nautical terms. 
In future work, we plan to experiment with contextual embeddings and eventually integrate them into our NER service. 

\paragraph{Acknowledgments:} This work was funded by the \textit{Volkswagen Foundation} under Grant No. 90~847, and by the \textit{DAAD} via a WISE stipend. 

\bibliographystyle{konvens2018}
\bibliography{references}

\end{document}